\pdfoutput=1
\documentclass[11pt]{article}
\usepackage{booktabs} % 在导言区加载booktabs宏包
\usepackage{float}

\usepackage{acl}
\usepackage{longtable}
\usepackage{times}
\usepackage{latexsym}
\usepackage{pgffor}
\usepackage{caption} % 用于minipage中的独立标题
\usepackage{subcaption} % 用于minipage中的子标题，如果需要的话

\usepackage{caption}
\usepackage{geometry} % 可以调整页面边距，确保内容不会超出页面
 
\usepackage{tabularx}
\usepackage{array}

\usepackage[T1]{fontenc}
\usepackage[utf8]{inputenc}

\usepackage{microtype}

\usepackage{inconsolata}

\usepackage{algorithmic}
\usepackage{caption}
\usepackage{graphicx}
\usepackage{textcomp}
\usepackage{xcolor}
\usepackage{tabularx}
\usepackage{booktabs}
\usepackage{makecell}
\usepackage{hyperref}
\usepackage{multirow}
\usepackage[normalem]{ulem}
\useunder{\uline}{\ul}{}
\usepackage{threeparttable}
\usepackage{listings}
\usepackage{amsmath}

\usepackage[ruled,linesnumbered]{algorithm2e}

\usepackage{cleveref}
\title{SilverSight: A Multi-Task Chinese Financial Large Language Model Based on Adaptive Semantic Space Learning}

\author{
\textbf{Yuhang Zhou}\thanks{Email: yuhangzhou22@m.fudan.edu.cn}\(^{1,2}\) \quad
\textbf{Zeping Li}\(^{1,2}\) \quad
\textbf{Siyu Tian}\(^{1,2}\) \quad
\textbf{Yuchen Ni}\(^{3}\) \\
\textbf{Sen Liu}\(^{1,2}\) \quad
\textbf{Guangnan Ye}\thanks{Corresponding Author. Email: yegn@fudan.edu.cn}\(^{1,2}\) \quad
\textbf{Hongfeng Chai}\(^{1,2}\) \\
\\
\(^{1}\)Institute of Fintech, Fudan University \\
\(^{2}\)School of Computer Science, Fudan University \\
\(^{3}\)School of Electronics and Information Engineering, Tongji University \\
}

\begin{document}
\maketitle

\begin{abstract}
Large language models (LLMs) are increasingly being applied across various specialized fields, leveraging their extensive knowledge to empower a multitude of scenarios within these domains. However, each field encompasses a variety of specific tasks that require learning, and the diverse, heterogeneous data across these domains can lead to conflicts during model task transfer. In response to this challenge, our study introduces an Adaptive Semantic Space Learning (ASSL) framework, which utilizes the adaptive reorganization of data distributions within the semantic space to enhance the performance and selection efficacy of multi-expert models. Utilizing this framework, we trained a financial multi-task LLM named "SilverSight". Our research findings demonstrate that our framework can achieve results close to those obtained with full data training using only 10\% of the data, while also exhibiting strong generalization capabilities.
\end{abstract}
% \textit{"Be fearful when others are greedy and to be greedy only when others are fearful."}
% \begin{flushright}
% --Warren Buffett
% \end{flushright}

\section{Introduction}\label{sec1}
Recently, large language models such as GPT-4~\cite{openai2023gpt4} and LLaMA~\cite{touvron2023llama} have demonstrated remarkable capabilities across various tasks in the field of natural language processing (NLP). These models, with their formidable knowledge storage and context comprehension abilities, have extensively infiltrated professional domains such as finance~\cite{zhou2024large,chen2023disc} and law~\cite{cui2023chatlaw}, showcasing their exceptional performance. Instruction fine-tuning, leveraging supervised datasets to refine the models, plays a important role in the construction of LLMs, facilitating their transition from simple text continuation to solving complex tasks~\cite{qiu2020pre}. However, the process is often constrained by the data availability in specific domains and computational resource limitations, necessitating the use of parameter-efficient fine-tuning 
(PEFT) strategies~\cite{houlsby2019parameter}. These strategies, requiring minimal updates or additions to the base model's parameters, significantly enhance the model's responsiveness to instructions. In this context, the Low-Rank Adaptation (LoRA) method~\cite{hu2021lora} emerged, introducing decomposable low-parameter side matrices to the original models. This approach allows for modular modifications and comprehensive performance improvements while offering the convenience of a "plug-and-play" feature.

\begin{figure*}[htbp]
\centerline{\includegraphics[width=0.9\textwidth]{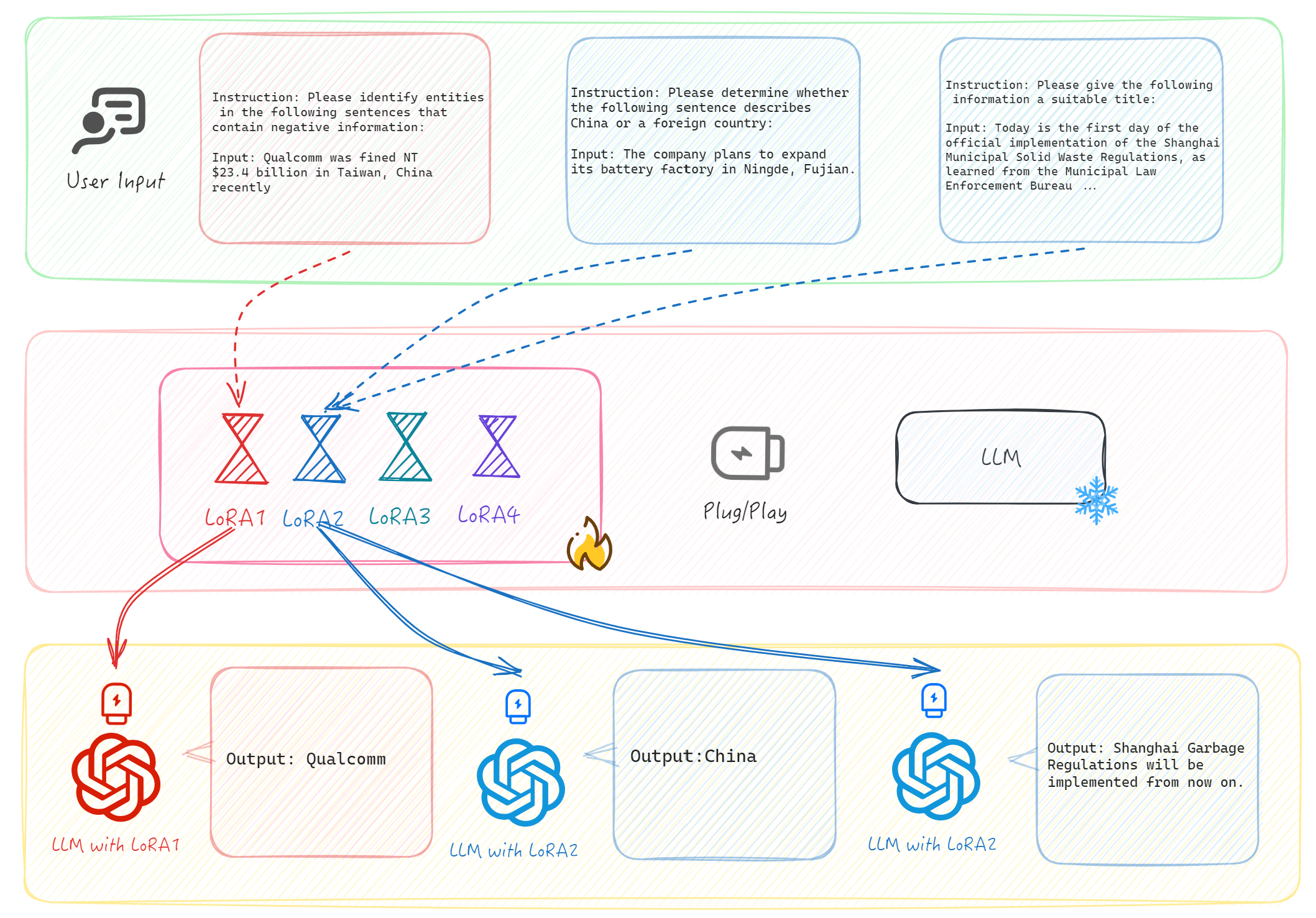}}
\caption{Schematic Diagram of the SilverSight Framework: when the user inputs a financial question, select the most suitable expert to answer the question.}
\label{demo}
\end{figure*}

Currently, one of the research hotspots in the field of NLP is the use of Mixture of Experts (MoE) models~\cite{jordan1994hierarchical} to address the challenges of multitasking. MoE models, an ensemble learning strategy, integrate multiple expert models to leverage each model's expertise on specific sub-problems, thereby enhancing overall performance in complex tasks~\cite{dou2023loramoe}. Zadouri~\cite{zadouri2023pushing} initially proposed combining the LoRA approach with the MoE framework, employing a token-level soft routing mechanism to weight and integrate the outputs of various expert models, effectively improving model performance. Subsequently, researchers introduced the LoRAMoE method~\cite{dou2023loramoe}, which divides expert models into two groups: one focused on processing general world knowledge, and the other dedicated to new tasks encountered during instruction fine-tuning. This design aims to bolster the model's capability in downstream tasks while retaining the accumulated world knowledge in models. While these multi-expert strategies demonstrate advantages in addressing diverse issues, their token-level processing and the fixed nature of post-training experts and routing strategies limit the models' ability to rapidly adapt to new instructions or scenarios. To address this, some researchers proposed the LoraRetriever method~\cite{zhao2024loraretriever}, drawing from sentence-level retrieval concepts. It uses the average of 12 data embeddings in each expert model as LoRA's inherent embeddings and leverages the embedding of the input question to retrieve the most matching LoRA, thereby enhancing the performance and generalization ability of the hybrid training model.

Although these methods excel in enhancing the dynamic adaptability of adapters like LoRA, they often rely on manually segmenting training data based on task types during LoRA training, without fully considering the intrinsic connections within the semantic space of the data. This approach may lead to biased embeddings in LoRA experts, thereby affecting the overall system performance. Meanwhile, data from different sources vary in diversity and quality~\cite{touvron2023llama}, and the appropriate configuration of different data volumes has not been thoroughly explored in previous research. This exacerbates the issues of LoRA adapter and data selection, posing a key challenge that needs addressing. Therefore, exploring a method that can optimize both LoRA selection and data redistribution strategies to maintain continuity and connectivity of data in the semantic space becomes a crucial research direction for enhancing the performance of hybrid expert models. This requires focusing not only on the training process of LoRA but also on data preprocessing and selection mechanisms, ensuring the model maintains efficient and accurate performance when faced with complex and varied tasks.

To address the gap, we propose an Adaptive Semantic Space Learning (ASSL) framework and have trained a financial multitask large language model named "SilverSight" using a Chinese financial multitask dataset. The ASSL framework clusters multitask training data based on similarities in the semantic space, thereby eliminating the need for predefined task types. Our experiments show that our method has clear advantages over traditional predefined classification methods, ensuring that each expert model is allocated to the most relevant downstream tasks. We also observed that training tasks from different sources and formats could lead to unavoidable task conflicts, affecting the model's ability to follow instructions. Moreover, by leveraging similarities in the semantic space, our method can aggregate diverse data from complementary tasks, improving model performance on related tasks. Within the ASSL framework, we adopt a model self-evolution-based data redistribution strategy in each cluster for adaptive data selection. Through a two-step adaptive data filtering process, we ensure that the limited data used to train each LoRA expert is of high quality, coverage, and necessity. In scenarios where original data quality and quantity distributions are uneven, the ASSL framework adapts the selection of training data for each LoRA by considering semantic space distribution density and model self-feedback mechanisms. This approach allows the model to more evenly fit top similarity data and tail small-sample data in the long-tail distribution, thereby improving the model's generalization ability and performance across diverse tasks.

We collected 220k Chinese financial fine-tuning data points from 23 different sources in the financial domain, classifying them into various task types such as sentiment analysis, financial Q\&A, text generation, and financial multiple-choice questions. Based on these experimental data, we made the following key findings:

(1)By clustering based on similarities in the semantic space, we could identify mutually enhancing and conflicting training tasks. Using multiple expert models to learn specific domain tasks allows each model to focus on its area of expertise, achieving a division of labor.

(2)By combining the density distribution of data in the semantic space with the model's own training data needs, we could effectively perform semantic smoothing and redistribution of data. This method enables the entire system to achieve similar effects with only 10\% of the data used for fine-tuning compared to using the full dataset.

(3)By smoothing the data distribution within clusters, we use the centroid of data embeddings within a cluster as the embedding for LoRA experts, optimizing the selection of LoRA.

These findings not only validate the effectiveness of the ASSL framework in financial multitask learning but also provide new perspectives and methods for efficiently using limited data resources to improve model performance. These achievements are significant for advancing natural language processing technology in the domain.

Our main contributions are as follows:

\begin{enumerate}
\item We innovatively propose the ASSL framework, which analyzes the distribution of data in the semantic space to achieve an adaptive selection mechanism for LoRA experts and data in multitask scenarios. This effectively adjusts the distribution balance between experts and training data, using semantic space information for smooth data processing, thereby avoiding expert overfitting or conflicts during training.
\item We empirically validate the effectiveness of the ASSL framework in the financial domain and successfully trained a Chinese financial multitask large language model, "SilverSight". We conducted comprehensive tests on the model's generalization ability and multitask performance on two evaluation benchmarks, fully demonstrating the framework's superiority and application potential.
\item We conducted an in-depth multidimensional analysis of the sources of different model capabilities, providing new directions for future research in the field of adaptive selection of experts and data.
\end{enumerate}

\section{Preliminaries}
\label{prelim}
\subsection{LoRA}
Fine-tuning LLMs with all parameters requires substantial data and computational resources and may lead to issues of catastrophic forgetting~\cite{luo2023empirical}. To address this, recent studies have developed various efficient fine-tuning methods, among which LoRA~\cite{hu2021lora} has been widely applied due to its effectiveness. The LoRA method employs the concept of low-rank matrix decomposition to reduce the number of parameters that need to be adjusted during fine-tuning. This is achieved by introducing an additional bypass structure \(W_0^{m \times n}\) in the network. Specifically, LoRA modifies the weights in the network by adding \(W_0^{m \times n}\) to the original weight matrix \(W^{m \times n}\), as follows:
\begin{equation}
\begin{split}
W'^{m \times n} &= W^{m \times n} + W_0^{m \times n} \\
&= W^{m \times n} + A^{m \times r}B^{r \times n}
\end{split}
\end{equation}

Here, \(A^{m \times r}\) and \(B^{r \times n}\) are two matrices obtained from \(W_0^{m \times n}\) through low-rank decomposition, with r representing the rank of the matrices. When the input to the model is the token representation \(x^{n \times 1}\), the output of the model \(y^{m \times 1}\) can be expressed as:
\begin{equation}
y = W'x = \left(W + AB\right)x = Wx + ABx
\end{equation}
In this case, \(x^{n \times 1}\) represents the token representation input to the model, while \(y^{m \times 1}\) denotes the model output.

\begin{figure*}[htbp]
\centerline{\includegraphics[width=1\textwidth]{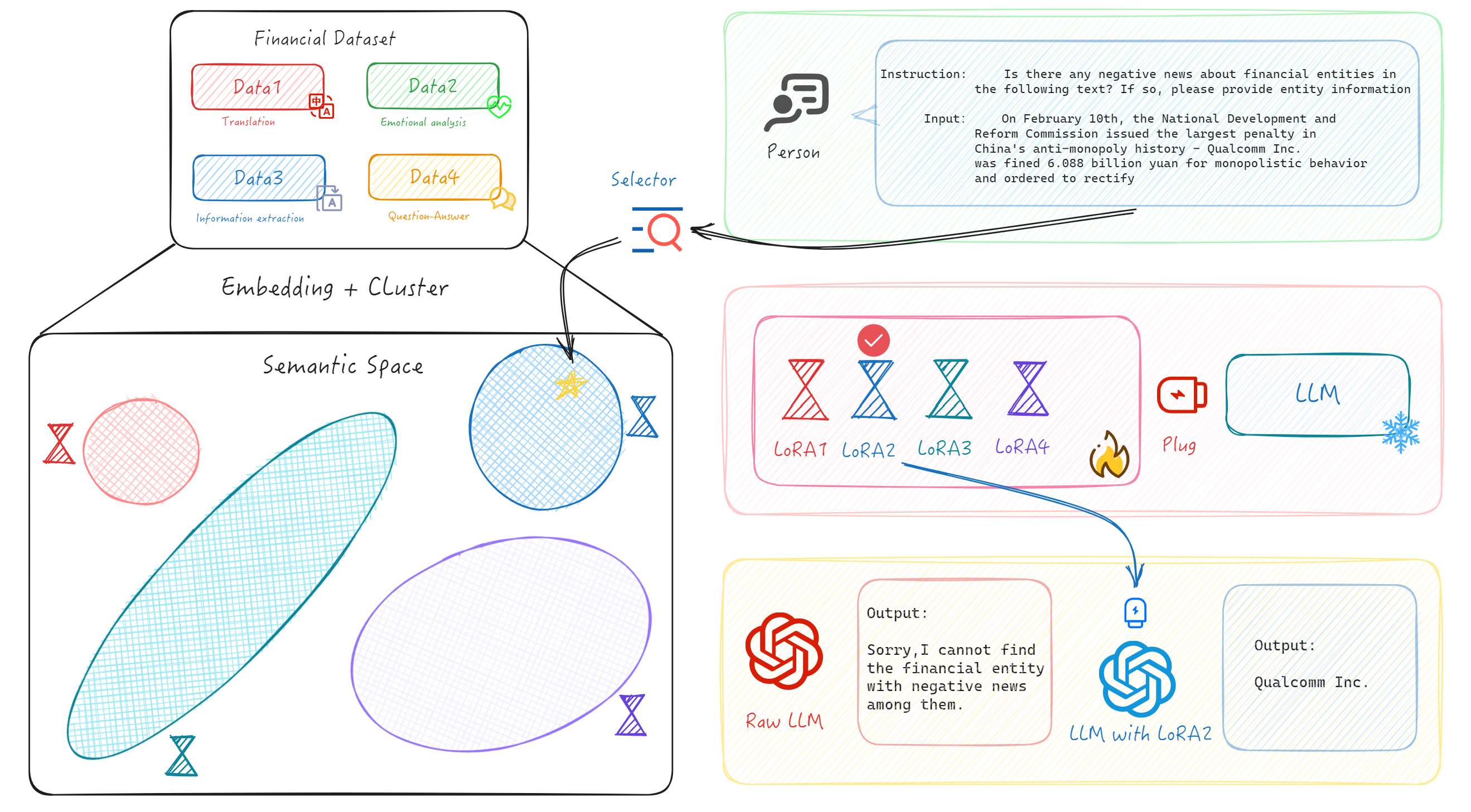}}
\caption{Adaptive Semantic Space Learning Framework.}
\label{fig:silversight}
\end{figure*}

\subsection{Task Definition}
Within the ASSL framework, we harness the distribution of data in the semantic space to design a multi-expert system, focusing primarily on two core tasks: adaptive expert selection and adaptive data selection. These tasks aim to enhance the system's flexibility and efficiency in a multi-task environment.

The task of adaptive expert selection concentrates on how to adaptively choose the most suitable LoRA expert for a specific task given an input. Considering that our multi-task system comprises a set of \( n \) different tasks \( T = \{t_1, t_2, \ldots, t_n\} \) and a set of \( m \) different LoRA experts \( E = \{e_1, e_2, \ldots, e_m\} \), where each expert \( e_i \) is trained on several specific tasks to optimize its performance. For any input \( x \), our goal is to find a mapping function \( f: X \to E \) such that \( f(x) \) selects the expert \( e \) most suited for the current input \( x \). This can be achieved by maximizing the relevance between input \( x \) and expert \( e \), as follows:

\begin{equation}
    e^* = f(x) = \underset{e \in E}{\arg\max}\; \text{Relevance}(x, e)
\end{equation}

Here, the function \( \text{Relevance}(x, e) \) measures the relevance between the input \( x \) and the expert \( e \).

On the other hand, the task of adaptive data selection focuses on handling heterogeneous data from multiple sources and addressing the long-tail distribution present within the data. We have a dataset \( D = \{d_1, d_2, \ldots, d_l\} \), with each \( d_i \) representing data from different sources. In practice, these data often follow a long-tail distribution, meaning a large amount of data is concentrated in a few categories, while most categories contain only a small amount of data. Our goal is to process the original dataset \( D \) through a transformation function \( g: D \to D' \), converting it into a smoother, more evenly distributed dataset \( D' \) to mitigate the effects of the long-tail distribution.

\section{Adaptive Semantic Space Learning}
\label{sec3}

We introduce a framework named Adaptive Semantic Space Learning (ASSL), designed to facilitate the adaptive selection of LoRA experts and their corresponding data. The framework is illustrated in~\ref{fig:silversight}, ensuring that each expert operates at peak performance and that the system excels across a variety of multitask environments. In this chapter, we will delve into the framework's two pivotal components, detailing how adaptive selection between experts and data is achieved through the redistribution of data within the semantic space.

\subsection{Adaptive Selection of LoRA Experts}
In addressing the adaptive selection of LoRA experts, our aim is to optimize data segmentation to avert potential conflicts between tasks and ensure that for each input, the expert most proficient in handling the given problem is matched. This objective is twofold: (1) how to more effectively train multitask instructions; and (2) how to choose the most appropriate LoRA expert based on user input.

First, to enhance the system's generalization capability across diverse instructions, we expanded the instructions for each subtask within the task set. Specifically, we manually crafted 30 semantically similar but differently phrased instructions for each task. Then, using a sentence encoder $\text{Emb}(\cdot)$, we encoded each instruction and its concatenated input data \( \text{Ins} \oplus \text{Inp} \), obtaining embedding vectors for all data within the same semantic space. We employed the K-means clustering method~\cite{macqueen1967some} to group semantically similar sentences in the semantic space into \( K \) clusters, thereby optimizing task data segmentation. The clustering process is represented by the following equation:

\begin{equation}
    \text{Cluster}_k = \underset{x \in X}{\arg\min} \sum_{i=1}^{K} \sum_{x \in C_i} \lVert x - \mu_i \rVert^2
\end{equation}

where \( X \) denotes the set of all data points, \( C_i \) is the set of data points in cluster \( i \), and \( \mu_i \) is the centroid of cluster \( i \).

As proven by experiments in Section~\ref{sec:experience.exablation}, compared to mixed task segmentation and predefined label-based segmentation, this semantic clustering approach significantly enhances system performance. For each LoRA expert, we selected the centroid of its cluster as the expert's semantic embedding. The centroid \( \mu_i \) for each cluster is the average position of all semantic embeddings in the cluster, calculated as follows:

\begin{equation}
   \text{Emb}(e_i) = \mu_i = \frac{1}{|C_i|} \sum_{x \in C_i} x
\end{equation}

where \( C_i \) is the set of data points in cluster \( i \) and \( |C_i| \) represents the number of elements in set \( C_i \). \( \mu_i \) signifies the mean of all points in cluster \( i \), i.e., the centroid. Whenever there is a user input, the system finds the expert whose semantic embedding is closest to the semantic embedding of the user input using the following formula:

\begin{equation}
    e^* = \underset{e \in E}{\arg\min} \lVert \text{Emb}(x) - \text{Emb}(e) \rVert
\end{equation}

Here, \( \text{Emb}(x) \) is the embedding vector of the user input, \( \text{Emb}(e) \) is the semantic embedding vector of the expert, and \( e^* \) is the selected expert. Through this matching method, the system can find the expert that best matches the training task in the semantic space.

\begin{figure*}[htbp]
\centerline{\includegraphics[width=1\textwidth]{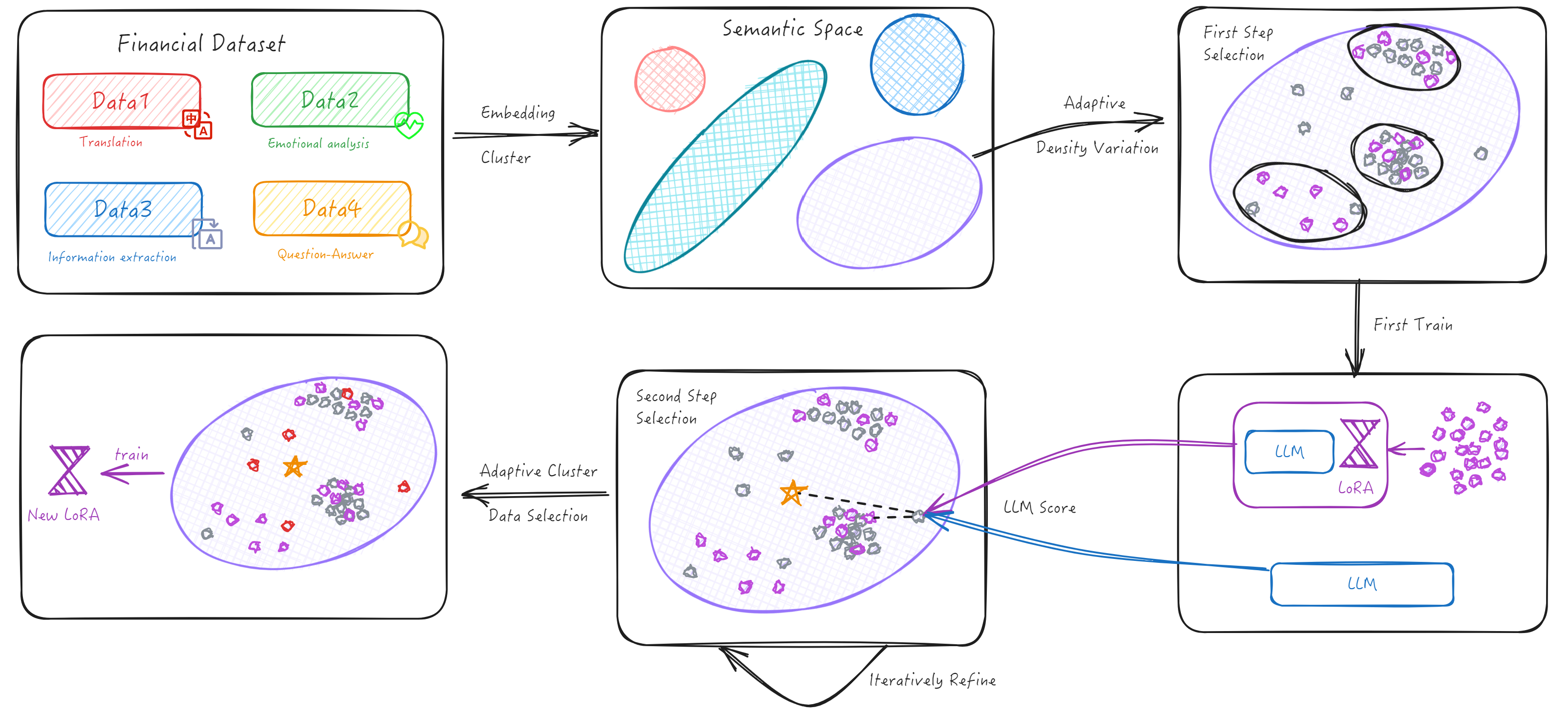}}
\caption{Adaptive Semantic Space Data Redistribution Process.}
\label{fig:data}
\end{figure*}

\subsection{Adaptive Selection of Heterogeneous Multi-source Data}

In this section, we discuss clustering multi-task supervised data to isolate conflicting tasks and gather those that enhance each other. While this approach effectively resolves conflicts between tasks, it introduces new challenges such as imbalanced data ratios and varying quality, especially concerning long-tail data distributions. To address these issues, we designed a two-stage data redistribution operation for each cluster, aiming to achieve effective fine-tuning on a limited dataset while emulating the effects of full dataset fine-tuning, as shown in~\autoref{fig:data}.

In the first stage, addressing the issue of data imbalance within clusters, we devised an adaptively adjusted A-DBSCAN algorithm based on the DBSCAN algorithm~\cite{schubert2017dbscan}, performing nested clustering within each cluster. The detailed algorithmic process is provided in Appendix~\ref{appendix:DBSCAN}. This algorithm dynamically adjusts the requirements for connectivity count based on data density in different areas. The specific steps are as follows: initially, the algorithm evaluates the local density of data points in the semantic space using the K-Nearest Neighbor (KNN)~\cite{peterson2009k} distance calculation framework. The local density for each data point is the reciprocal of the average distance to its \(k\) nearest neighbors, mathematically expressed as:

\begin{equation}
    \rho_i = \frac{1}{\frac{1}{k} \sum_{j=1}^{k} d(x_i, x_{ij})}
\end{equation}

where \(d(x_i, x_{ij})\) represents the distance between data point \(x_i\) and its \(j\)th nearest neighbor \(x_{ij}\).

Subsequently, the algorithm sorts data points based on the calculated local density values, forming a priority queue and prioritizing data points with higher local density. In each iteration, the point with the highest local density in the queue is selected as the starting point for cluster formation.

During the adaptive process, the algorithm defines the neighborhood radius \(\varepsilon\) using the median of KNN distances for all data points in the queue. The initial value of neighborhood node count \(MinPts\) is heuristically set as:

\begin{equation}
    MinPts_{\text{init}} = \frac{\varepsilon \times \rho_{\text{max}}}{2}
\end{equation}

where \(\rho_{\text{max}}\) denotes the global maximum local density, i.e., the local density of the first data point in the initial priority queue. After forming each cluster, \(MinPts\) is updated according to the following formula to adapt to the current local density environment:

\begin{equation}
    MinPts_{\text{update}} = \max\left(2, \frac{\rho_{\text{current}}}{\rho_{\text{max}}} \times MinPts\right)
\end{equation}

Here, \(\rho_{\text{current}}\) refers to the local density of the first point in the current priority queue. This dynamic adjustment strategy allows the algorithm to adapt more flexibly to different density data distributions, enhancing clustering accuracy and efficiency. Simultaneously, this strategy enables averaging data selection in each sub-cluster while filtering out disconnected noise points, effectively downsampling high-density data and upsampling low-density data to avoid overfitting and underfitting.

In the second stage, we use the small dataset filtered from the first stage to conduct preliminary fine-tuning on the model. According to previous studies, large language models primarily learn new language style distributions during the fine-tuning phase and struggle to acquire new domain knowledge. Therefore, considering the score differences before and after training the large model on unselected data, we designed two scoring mechanisms to evaluate each data point's value to the current model. The differential score and proportional score are defined as follows:

\begin{equation}
    \text{Score}_{\text{diff}}(x) = \text{LLM}_{\text{Raw}}(x) - \text{LLM}_{\text{LoRA}}(x)
\end{equation}

\begin{equation}
    \text{Score}_{\text{prop}}(x) = -\frac{\text{LLM}_{\text{LoRA}}(x)}{\text{LLM}_{\text{Raw}}(x)+1}
\end{equation}

\begin{equation}
    \text{Score}_{\text{llm}}(x) = \text{Score}_{\text{diff}}(x)+\text{Score}_{\text{prop}}(x)
\end{equation}

Model scores are calculated using the Rouge method~\cite{lin2004rouge}. Moreover, to ensure the quality of newly selected data and coverage over the

 entire cluster, we introduce and modify the MMR formula~\cite{carbonell1998use} as the utility function for data:

\begin{equation}
\begin{split}
\text{U}(d_{\text{new}}) &= \lambda_1 \cdot \text{sim}(\mu, d_{\text{new}}) \\
&- \lambda_2 \cdot \max_{d \in D_{\text{selected}}} \text{sim}(d, d_{\text{new}}) \\
&+ \lambda_3 \cdot \text{Score}_{\text{llm}}(d_{\text{new}})
\end{split}
\end{equation}

where \(\mu\) represents the cluster centroid, \(d_{\text{new}}\) is the new data point to be added, \(D_{\text{selected}}\) denotes the set of already selected data points, and \(\lambda_1\), \(\lambda_2\), and \(\lambda_3\) are three adjustable weight parameters used to balance the contributions of similarity, diversity, and model score to the final utility value of data points.

Through two stages of data filtering, data within each cluster will be redistributed in the semantic space, encouraging the model to learn rare but beneficial data and avoid overfitting on common datasets.

\section{Experiments}
\label{sec:experience}
In this chapter, we will train a Chinese financial multitask large model named "SilverSight" using publicly available datasets from the financial domain. We will validate the effectiveness of the Adaptive Semantic Space Learning (ASSL) algorithm proposed in this study using two Chinese financial evaluation benchmarks.

\subsection{Data Introduction}
We collected 220,000 Chinese financial fine-tuning data points from 23 different sources within the financial domain, classifying these data into 7 task categories such as sentiment analysis and financial Q\&A. Detailed information can be found in Appendix~\ref{appendix:FinNLP}. For evaluation, we used the CFLEB~\cite{lu2023bbt} and FinEval~\cite{zhang2023fineval} benchmarks, aiming to assess the large language model's knowledge reserve, instruction following, and task execution capabilities in the financial domain.

The CFLEB benchmark, constructed using publicly available research reports and news items, is a high-quality and practical evaluation standard containing six natural language processing tasks. It measures the comprehensive understanding and generation capabilities of models in the financial domain across sentiment analysis, Q\&A, summary generation, information extraction, semantic matching, and more.

The FinEval benchmark is a comprehensive dataset designed for knowledge assessment in the financial domain, consisting of 4661 multiple-choice questions covering 34 different academic fields such as finance, insurance, macroeconomics, and tax law. The questions are mainly divided into four categories: finance, economics, accounting, and qualification exams, to thoroughly test large models' general knowledge in the financial domain and evaluate their advanced knowledge and practical application capabilities in financial specialties.

\subsection{Experimental Setup}
We used the representative Chinese model Qwen1.5-7B-Chat~\cite{bai2023qwen} as our base model and conducted our experiments on 2 A800 GPUs. For each expert, we trained a LoRA adapter with \(r\) set to 16 and \(\alpha\) set to 32. For all fine-tuning, we set the learning rate to 1e-4, with a warm-up over 10\% of steps, followed by a cosine decay to 1e-5, over a total of 3 training epochs. In the K-Means clustering algorithm, after multiple trials, we finally set \(K=6\), and in the adaptive A-DBSCAN algorithm, we chose \(k=20\) as the initial KNN calculation parameter. For the modified MMR formula, we set \(\lambda_1\), \(\lambda_2\), and \(\lambda_3\) to 0.2, 0.2, and 0.6, respectively, to balance the contributions of similarity to the cluster center, diversity, and model score to the final utility function of the data points.

\subsection{SilverSight: Financial Multitask Large Model}
In this study, we processed data using the designed ASSL framework and trained a multitask large language model for the Chinese financial domain named "SilverSight". First, we employed the K-Means algorithm to cluster publicly available financial domain data in the semantic space, ultimately forming six categories. The data distribution for various category numbers can be found in Appendix~\ref{app:data_mix}. Next, we smoothed the data distribution within each cluster through two stages: In the first stage, we selected data using the adaptive density clustering algorithm A-DBSCAN, downsampling high-density areas and upsampling low-density areas within each cluster, selecting approximately 2000 fine-tuning data points per cluster and treating data in extremely low-density areas as noise. The second stage aimed to adaptively supplement necessary data points not selected in the first stage, using the original model before training and the model after training with the first-stage data. This made the fine-tuning data distribution smoother and enhanced data selection diversity. In total, about 4000 data points were selected from each cluster, amounting to approximately 10\% of the total data volume. We trained six different LoRA expert models with these six adaptively semantically smoothed data categories to adapt to different tasks in the financial domain. When addressing specific financial issues, the most suitable LoRA expert is automatically selected by calculating the similarity between the problem's representation in the same semantic space and the representations of the six LoRA experts, achieving an efficient, adaptive, and well-coordinated financial multitask large model system.

\begin{table*}[htbp]

    \centering
    \caption{Evaluation results of CFLEB: All-data represents a single model trained using all data we collected, Cluster-data represents a multi expert system trained using all clustering data, and SilverSight-mix represents a single model trained using all filtered data. In the evaluation results, our SilverSight model trained with 10\% data achieved similar results to All-data, while the Cluster-data system achieved the best performance.}
    \label{cfleb_experiment}
    \begin{tabular}{ccccccccc}
        \hline
        \textbf{Model} & \textbf{FE} & \textbf{QA} & \textbf{NA} & \textbf{RE} & \textbf{FinNSP1} & \textbf{FinNSP2} & \textbf{NL} & \textbf{AVG} \\
         & (ACC) & (F1) & (Rouge) & (F1) & (ACC) & (F1) & (F1) &  \\
        \hline
        Qwen-1.5-7B & 65.3 & 31.6 & 30.9 & 10.4 & 90.8 & 9.1 & 33.4 & 38.8 \\
        All-data & \textbf{72.3} & \textbf{87.1} & \textbf{53.8} & \textbf{35.8} & 93.2 & 31.1 & \textbf{90.1} & 66.2 \\
        Cluster-data & 70.9 & 83.4 & 53.1 & 35.3 & \textbf{93.5} & \textbf{56.2} & \textbf{90.1} & \textbf{68.9} \\

        \hline
        $SliverSight_{LoRA-0}$ & 68.2 & 60.8 & 32.5 & 10.9 & 83.8 & 10.6 & 45.7 & 44.6 \\
        $SliverSight_{LoRA-1}$ & \textbf{70.2} & \textbf{74.5} & 49.5 & \textbf{31.4} & 87.8 & \textbf{64.8} & 62.9 & \textbf{63.0} \\
        $SliverSight_{LoRA-2}$ & 65.4 & 61.9 & 45.2 & 11.9 & 11.8 & 45.4 & 25.9 & 38.2 \\
        $SliverSight_{LoRA-3}$ & 67.3 & 53 & 37.7 & 12.4 & \textbf{92} & 9.1 & 43.8 & 45.0 \\
        $SliverSight_{LoRA-4}$ & 57.5 & 72.9 & \textbf{49.6} & 14.9 & 89.2 & 18 & \textbf{80.5} & 54.7 \\
        $SliverSight_{LoRA-5}$ & 66.6 & 61.8 & 38.4 & 14 & 73.2 & 18.5 & 49.3 & 46.0 \\
        \hline
        SliverSight(our, 10\%) & \textbf{66.6} & 73.3 & \textbf{49.2} & \textbf{34.3} & \textbf{93.0} & \textbf{57.1} & \textbf{82.4} & \textbf{65.1} \\
        SliverSight-mix & 65.7 & \textbf{75.4} & 48.3 & 30.4 & 89.2 & 31.1 & 82.3 & 60.3 \\
        \hline
    \end{tabular}
\end{table*}

\subsection{Main Results}
The evaluation results of the "SilverSight" large model on CFLEB and FinEval benchmarks are shown in Tables~\ref{cfleb_experiment} and~\ref{FinEval_experiment}, respectively. In the CFLEB experiments, our approach achieved similar test scores to the fully fine-tuned model (Newest\_all) using only 10% of the total data and even significantly surpassed the full-data fine-tuned model on the FinNSP2 metric. Meanwhile, models fine-tuned on the full data set for each cluster showed improvements on the CFLEB dataset compared to the fully fine-tuned model, fully confirming the effectiveness of the ASSL framework. On the untrained FinEval benchmark, which tests the model's generalization ability and understanding of financial domain knowledge, the trained model showed a slight decrease in accuracy on math computation and qualification exam questions, but a noticeable increase in accuracy on economic and financial type questions, with an overall improvement on the FinEval average score compared to the original model.

\begin{figure}
    \centering
    \includegraphics[width=0.5\textwidth]{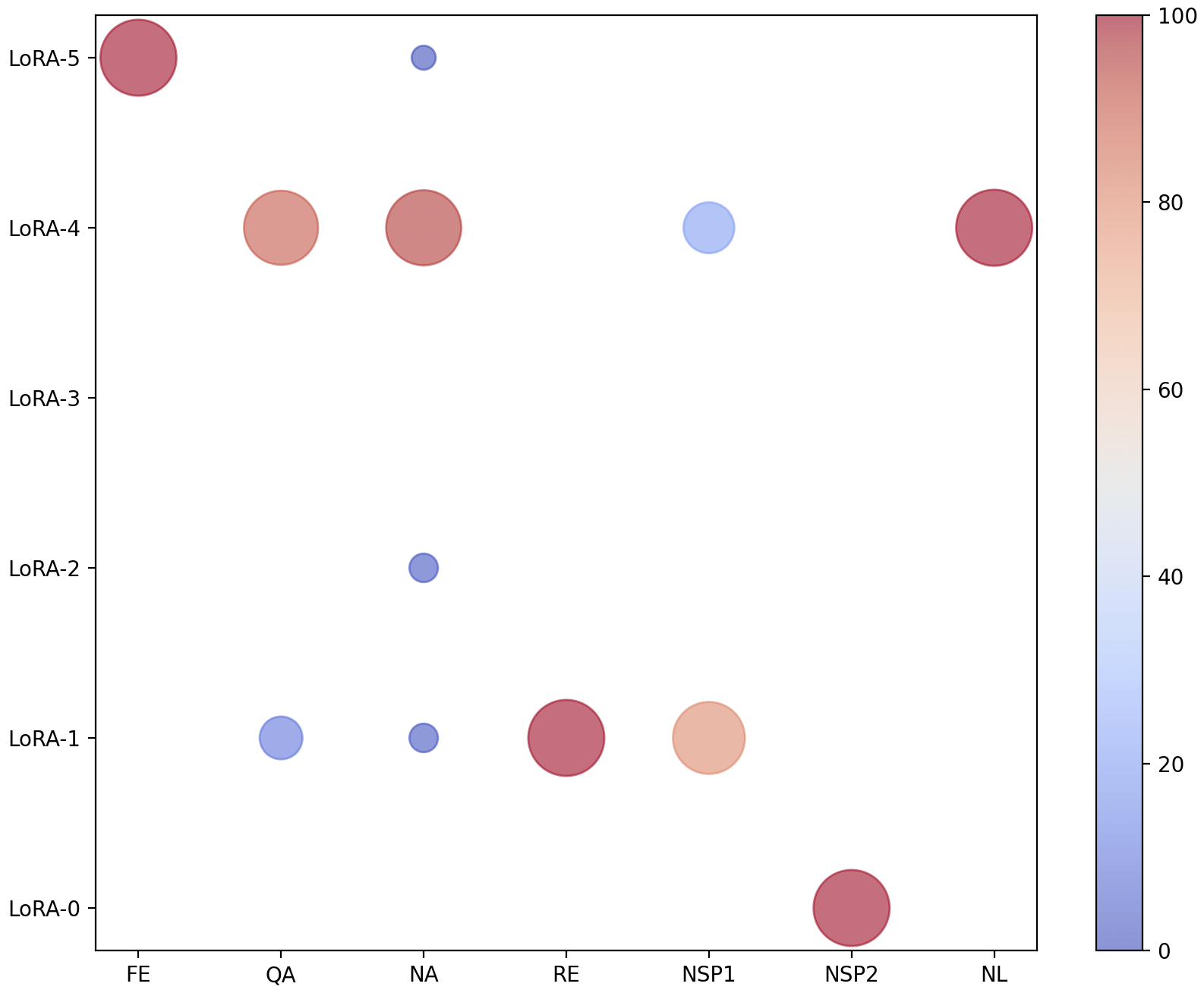}
    \caption{Expert Selection for Each Task Category on the CFLEB Dataset}
    \label{fig:bubble}
\end{figure}

% In the discussion above, we observed that using semantic-based clustering methods to classify data can significantly improve system performance compared to using manually assigned labels. After training multiple LoRA expert models with data from different categories, selecting the centroid of the LoRA expert model cluster as its semantic embedding and calculating its similarity with the question embedding can ensure the consistency and coherence of the semantic space, enhancing the system's overall performance.
To verify the effectiveness of the LoRA adaptive selection algorithm, we tested each of the six clustered LoRA experts individually on the CFLEB and FinEval datasets. The experimental results on these two evaluation datasets showed that the performance of the LoRA adaptive selection algorithm on each task was close to that of the single LoRA expert that performed best on that task, proving that most of the time, this adaptive selection algorithm could choose the most suitable LoRA expert to answer the question. Furthermore, as illustrated in~\autoref{fig:bubble}, the effectiveness of the LoRA adaptive selection algorithm is not only proven, but it is also demonstrated that using both clustering methods and the LoRA adaptive selection algorithm can ensure the consistency and coherence of the semantic space, ultimately enhancing the system's comprehensive performance.

\begin{table*}[h]
    \begin{center}
    \caption{Evaluation results of FinEval}
    \label{FinEval_experiment}
    \begin{tabular}{cccccc}
        \hline
        \textbf{FinEval} & \textbf{Accounting} & \textbf{Certificate} & \textbf{Economy} & \textbf{Finance} & \textbf{AVG} \\
        \hline
        Qwen-7B & 44.5 & 53.6 & 52.1 & 51.5 & 50.5 \\
        ChatGPT & 45.2 & 55.1 & 61.6 & 59.3 & 55.0 \\
        Qwen-1.5-7B & 69.5 & 71.3 & 62.8 & 65.6 & 67.8 \\
        GPT-4 & 59.3 & 70.4 & 74.5 & 71.0 & 68.6 \\
        \hline
        $SliverSight_{LoRA-0}$ & \textbf{68.9} & \textbf{72.2} & 64.7 & 65.9 & \textbf{68.3} \\
        $SliverSight_{LoRA-1}$ & 66.9 & 70.7 & 59.4 & 62 & 65.3 \\
        $SliverSight_{LoRA-2}$ & 62 & 65.6 & 58.5 & 62.6 & 62.6 \\
        $SliverSight_{LoRA-3}$ & 67.5 & 69.8 & \textbf{67.1} & 67.2 & 68 \\
        $SliverSight_{LoRA-4}$ & 68.5 & 70 & 63.8 & 66.2 & 67.5 \\
        $SliverSight_{LoRA-5}$ & 68.2 & 71.3 & 63.3 & 65.6 & 67.5 \\
        SliverSight(our, 10\%) & 67.9 & 70 & 66.7 & \textbf{67.9} & \textbf{68.3} \\
        \hline
    \end{tabular}
    \end{center}
\end{table*}

\subsection{Ablation Studies}
\label{sec:experience.exablation}

To demonstrate the effectiveness of the algorithms within the ASSL framework, we explored the following three questions:
\begin{enumerate}
\item Question 1: Does clustering in the semantic space have advantages over manual task categorization?
\item Question 2: Is the first stage of data redistribution meaningful?
\item Question 3: Is the second stage of data selection and supplementation meaningful?
\end{enumerate}

\begin{figure}
    \centering
    \includegraphics[width=0.4\textwidth]{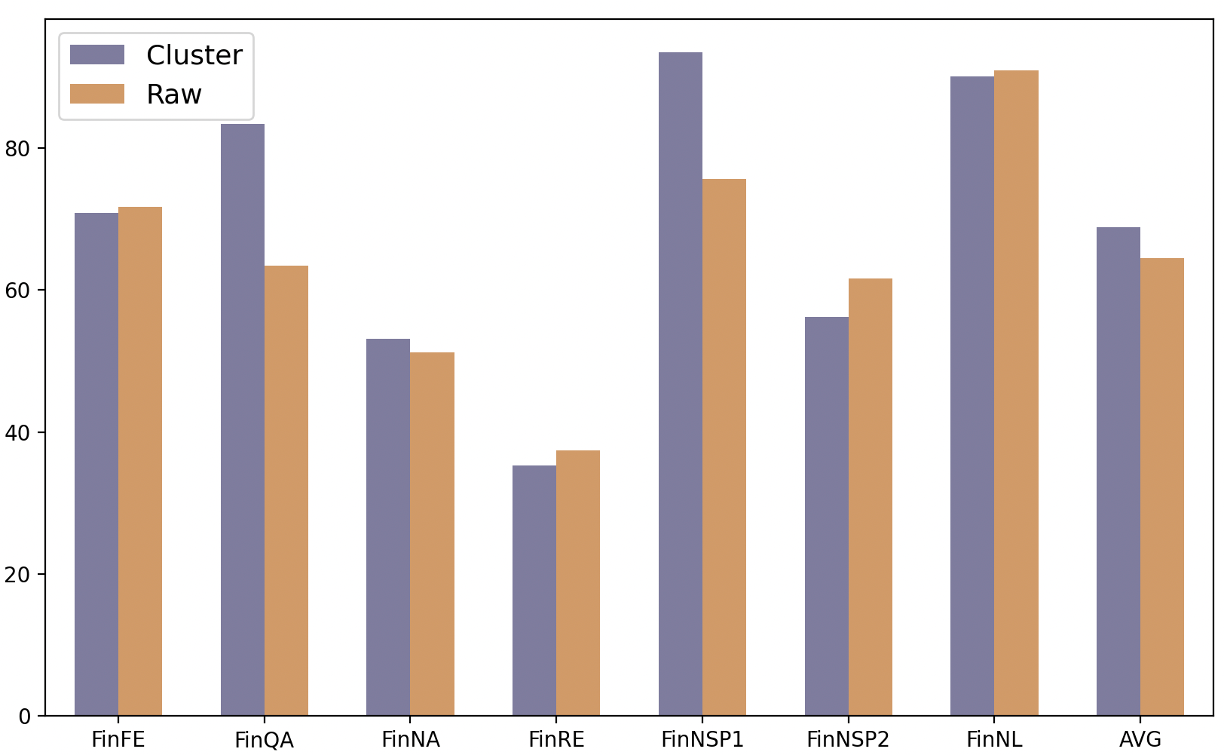}
    \caption{Evaluation Results of Clustering vs. Predefined Model Systems on the CFLEB Dataset}
    \label{Ablation1}
\end{figure}

To answer Question 1, we designed a set of control experiments with both model systems using the LoRA expert adaptive selection method. For the first model system, data was manually divided into seven categories based on the definitions of natural language processing tasks during the data preprocessing stage, with each category training a LoRA expert to form a multi-expert system. For the second model system, data was re-clustered into six categories in the semantic space to verify whether distances in the semantic space could reflect the complementarity between tasks, offering advantages over manually predefined categories. We conducted a thorough comparative analysis of the six categories obtained from clustering algorithms and the original seven categories. As shown in~\autoref{Ablation1}, the predefined manual categorization method performed worse on average on the CFLEB dataset compared to the semantic space clustering method, especially on the FinQA and FinNSP1 tasks. This proves the superiority of the semantic space clustering learning method over the predefined task type method, demonstrating that leveraging semantic space similarities can aggregate diverse data from complementary tasks, enhancing model performance on these tasks.

\begin{figure*}[htbp]
  \centering
  \begin{subfigure}{.5\textwidth}
    \centering
    \includegraphics[width=.9\linewidth]{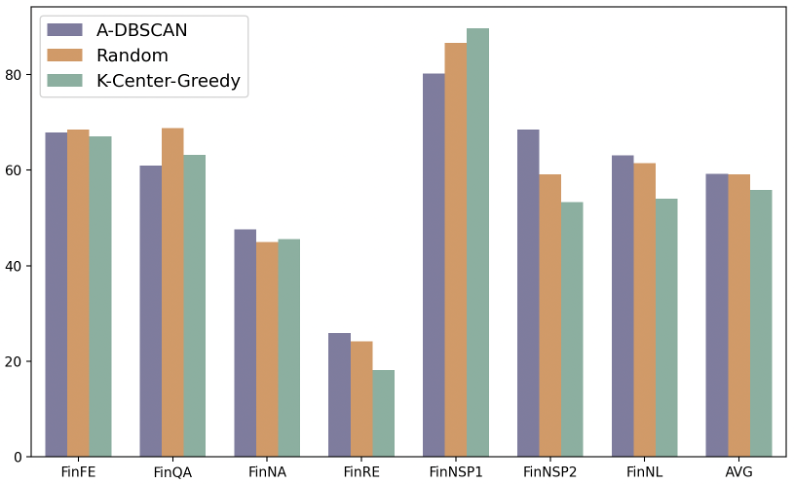}
    \caption{Evaluation results of CFLEB}
    \label{fig:sub1}
  \end{subfigure}%
  \begin{subfigure}{.5\textwidth}
    \centering
    \includegraphics[width=.9\linewidth]{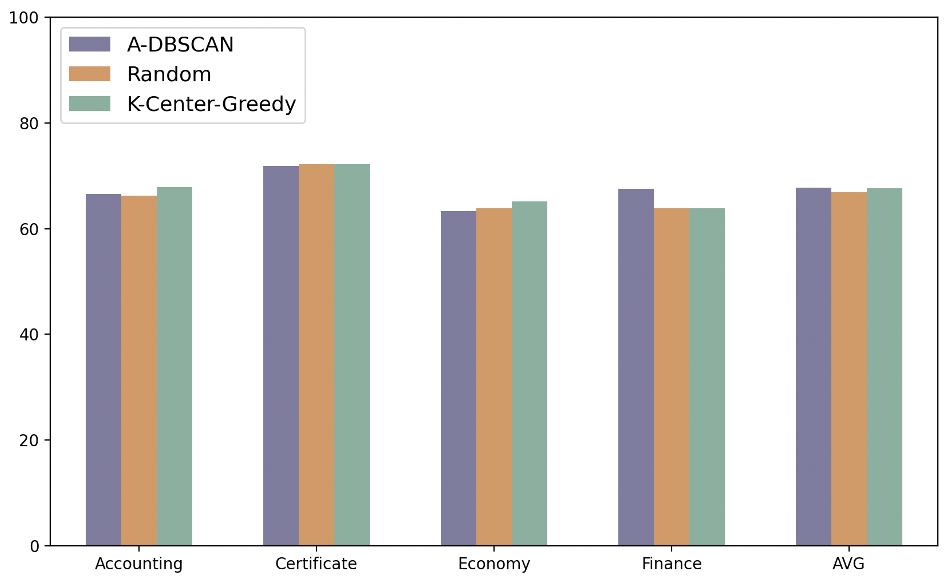}
    \caption{Evaluation results of FinEval}
    \label{fig:sub2}
  \end{subfigure}
  \caption{Ablation experiment 2: Comparison of data filtering algorithms in the first stage}
  \label{Ablation2}
\end{figure*}

\begin{figure*}[htbp]
  \centering
  \begin{subfigure}{.5\textwidth}
    \centering
    \includegraphics[width=.9\linewidth]{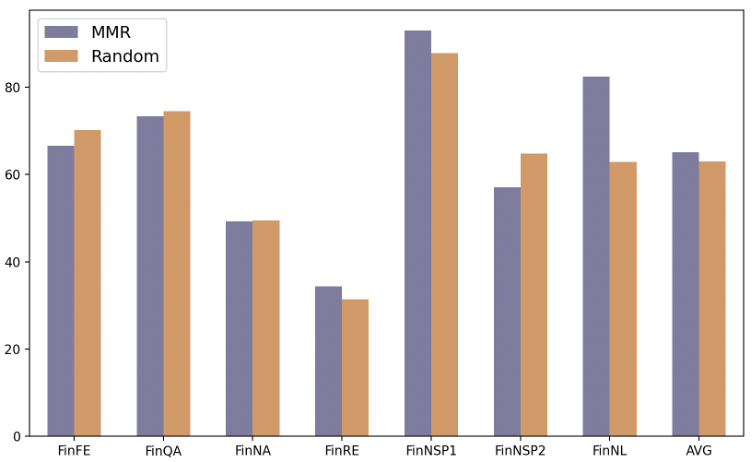}
    \caption{Evaluation results of CFLEB}
    \label{fig:sub1}
  \end{subfigure}%
  \begin{subfigure}{.5\textwidth}
    \centering
    \includegraphics[width=.9\linewidth]{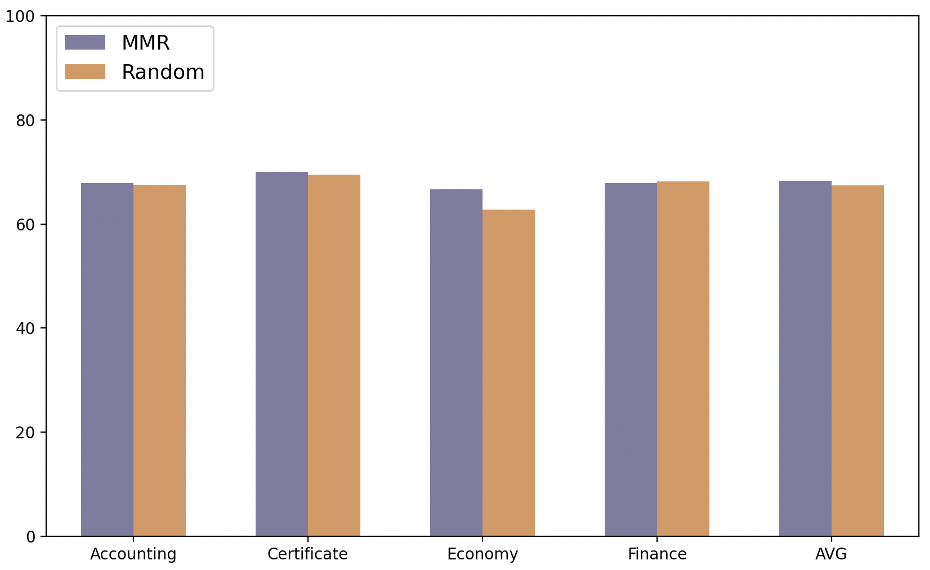}
    \caption{Evaluation results of FinEval}
    \label{fig:sub2}
  \end{subfigure}
  \caption{Ablation experiment 3: Comparison of data filtering algorithms in the second stage}
  \label{Ablation3}
\end{figure*}

\begin{table*}[htbp]
    \centering
    \caption{LoRA adaptive selection algorithm for CFLEB evaluation dataset}
    \label{Ablation4_1}
    \begin{tabular}{ccccccccc}
        \hline
        \textbf{Model} & \textbf{FE} & \textbf{QA} & \textbf{NA} & \textbf{RE} & \textbf{FinNSP1} & \textbf{NSP2} & \textbf{NL} & \textbf{AVG} \\
         & (ACC) & (F1) & (Rouge) & (F1) & (ACC) & (F1) & (F1) &  \\
        \hline
        MMR-0 & \textbf{68.2} & 60.8 & 32.5 & 10.9 & 83.8 & 10.6 & 45.7 & 44.6 \\
        MMR-1 & \textbf{70.2} & \textbf{74.5} & 49.5 & 31.4 & 87.8 & \textbf{64.8} & 62.9 & 63.0 \\
        MMR-2 & 65.4 & 61.9 & 45.2 & 11.9 & 11.8 & 45.4 & 25.9 & 38.2 \\
        MMR-3 & 67.3 & 53 & 37.7 & 12.4 & 92 & 9.1 & 43.8 & 45.0 \\
        MMR-4 & 57.5 & \textbf{72.9} & \textbf{49.6} & 14.9 & 89.2 & 18 & \textbf{80.5} & 54.7 \\
        MMR-5 & 66.6 & 61.8 & 38.4 & 14 & 73.2 & 18.5 & 49.3 & 46.0 \\
        MMR & 66.6 & \textbf{73.3} & \textbf{49.2} & \textbf{34.3} & \textbf{93.0} & \textbf{57.1} & \textbf{82.4} & \textbf{65.1} \\
        \hline
    \end{tabular}
\end{table*}

\begin{table*}[htbp]
    \begin{center}
    \caption{LoRA adaptive selection algorithm for FinEval evaluation dataset}
    \label{Ablation4_2}
    \begin{tabular}{cccccc}
        \hline
        \textbf{FinEval} & \textbf{Accounting} & \textbf{Certificate} & \textbf{Economy} & \textbf{Finance} & \textbf{AVG} \\
        \hline
        MMR-0 & \textbf{68.9} & \textbf{72.2} & 64.7 & 65.9 & \textbf{68.3} \\
        MMR-1 & 66.9 & 70.7 & 59.4 & 62 & 65.3 \\
        MMR-2 & 62 & 65.6 & 58.5 & 62.6 & 62.6 \\
        MMR-3 & 67.5 & 69.8 & \textbf{67.1} & \textbf{67.2} & 68 \\
        MMR-4 & 68.5 & 70 & 63.8 & 66.2 & 67.5 \\
        MMR-5 & 68.2 & 71.3 & 63.3 & 65.6 & 67.5 \\
        MMR-4k & 67.9 & 70 & \textbf{66.7} & \textbf{67.9} & \textbf{68.3} \\
        \hline
    \end{tabular}
    \end{center}
\end{table*}

To answer Question 2, we compared the first part of the ASSL framework's data redistribution A-DBSCAN algorithm with two baseline data selection algorithms: Random selection and K-Center Greedy algorithm. These three data selection algorithms have different focuses: the Random algorithm aims to uniformly extract data from each cluster according to the original data distribution, the K-Center Greedy algorithm focuses on the diversity of selected data, and the A-DBSCAN redistributes the original data based on density sparsity, also filtering a large amount of outlier noise data. As shown in~\autoref{Ablation2}, on the CFLEB dataset, the A-DBSCAN and Random algorithms showed prominent performance; whereas, on the FinEval dataset, the K-Center Greedy and A-DBSCAN algorithms demonstrated better performance. The robustness of the A-DBSCAN algorithm on both datasets confirms the superiority of this method, which smoothens the distribution of data in the semantic space, but due to its extensive noise filtering, further data supplementation is required.

To answer Question 3 and further confirm the importance of enhancing data smooth distribution, we compared the results of the second-stage improved MMR algorithm experiment with the Random algorithm. To ensure fairness, both experiment types selected 4000 data points from each cluster for training. As shown in~\autoref{Ablation3}, the comprehensive performance of the model significantly improved after the second-stage data expansion compared to just the first-stage filtering, and also outperformed the Random algorithm. This further validates the necessity and effectiveness of data expansion in the second stage, enhancing the smoothness of the original data's long-tail distribution.

\section{Discussion}
Our research trained a Chinese financial multitask large model based on the ASSL framework and achieved outstanding performance on multiple evaluation sets. The ASSL framework allows researchers to understand the advantages of adaptive learning based on semantic space, utilizing the characteristics of semantic space to infuse the benefits of a multi-expert system into the semantic space, achieving expert adaptive selection and data adaptive selection for better model performance. Based on this, we can further discuss the complementary characteristics of adaptive learning based on semantic space.

Each domain possesses numerous natural language processing or domain-specific tasks, requiring domain large models to have comprehensive and integrated capabilities. However, current heterogeneous multi-source data can easily trigger conflicts during task style transitions in models. When conducting multitask learning, training a large model with all corpora can easily affect the model's ability to follow instructions, leading to performance degradation and hallucination phenomena. Therefore, developing a multi-expert system and formulating adaptive task categorization and data selection strategies based on the distribution of all data in the same semantic space becomes particularly important. According to ~\autoref{fig:bubble}, utilizing the similarity of data points in the semantic space can effectively cluster task data that promote model capability enhancement and separate conflicting task data, improving the balance of fine-tuning data. Leveraging semantic space similarities not only optimizes the process of data categorization but also provides an efficient mechanism for selecting LoRA expert models. By comparing the similarity between the question embedding and the embeddings of various LoRA expert models, the system can accurately select the LoRA expert model that best matches the specific question. This method ensures the coherence and consistency of data, tasks, and models in the semantic space, ultimately enhancing the model's overall capabilities. Experimental results further validate the effectiveness and robustness of this method. Compared to randomly selecting data for model fine-tuning, using data selection algorithms can more accurately match the appropriate LoRA expert models, significantly improving the model's overall performance in handling different domain problems.

\section{Conclusion}
In this study, we mainly investigated the impact of adaptive learning based on semantic space on the performance of multi-expert large language model systems. The goal was to divide complementary and conflicting task data based on semantic space, adaptively select multiple experts using their embedding positions, and adjust the training data distribution in two stages using both the model itself and the density distribution of the semantic space. This framework enabled the model to have better performance and generalization when conducting multitask learning. Based on the Adaptive Semantic Space Learning framework, we trained the "SilverSight" large language model using publicly available datasets from the financial domain and evaluated the "SilverSight" model system on benchmarks, demonstrating its outstanding performance. These results not only prove the feasibility and effectiveness of our method but also open new perspectives for the development of multi-expert system large language models in the future.

\clearpage
\bibliography{sample}

\onecolumn
\appendix
\section{Appendix}

\subsection{Financial NLP Tasks}
\label{appendix:FinNLP}
This section provide a more detailed introduction to the financial NLP datasets used to train our model. The dataset covers seven types of tasks, including \textbf{Sentiment Analysis}, \textbf{Information Extraction}, \textbf{Text Classification}, \textbf{Text Generation}, \textbf{Semantic Matching}, \textbf{Financial Question Answering}, and \textbf{Financial Examination Multiple-choice Questions}. The specific statistical data are presented in~\autoref{dataset_statistic}.
\begin{table}[h]
\caption{Dataset Statistics}
\label{dataset_statistic}
\resizebox{\columnwidth}{!}{%
\begin{tabular}{llll} \hline
Dataset                 & Main Task Type  & Subtask Type       & Quantity (entries)  \\ \hline
flare-zh-fe         & Sentiment Analysis    & Sentiment Analysis         & 2020   \\
flare-zh-stocka     & Sentiment Analysis    & Sentiment Analysis         & 1477   \\ 
flare-zh-stockb     & Sentiment Analysis    & Sentiment Analysis         & 1962   \\ 
CFSC-ABSA           & Sentiment Analysis    & Sentiment Analysis         & 33184  \\ 
ficuge-finnl        & Text Classification    & Financial News Classification       & 7071   \\ 
flare-zh-nl         & Text Classification    & Financial News Classification       & 884    \\ 
flare-zh-nl2        & Text Classification    & Financial News Classification       & 884    \\ 
flare-zh-nsp        & Text Classification    & Financial Negative News and Subject Determination & 500    \\ 
fincuge-finna       & Text Generation    & News Summary       & 28799  \\ 
flare-zh-na         & Text Generation    & News Summary       & 3600   \\ 
flare-zh-19ccks     & Information Extraction    & Event Type and Entity Recognition    & 2936   \\ 
flare-zh-20ccks     & Information Extraction    & Event Type and Entity Recognition    & 9159   \\ 
flare-zh-21ccks     & Information Extraction    & Event Type and Entity Recognition    & 1400   \\ 
flare-zh-22ccks     & Information Extraction    & Event Type and Entity Recognition    & 11829  \\ 
fincuge-finre       & Information Extraction    & Relationship Extraction         & 13486  \\ 
flare-zh-ner        & Information Extraction    & Named Entity Recognition       & 337    \\ 
flare-zh-corpus     & Semantic Matching    & Semantic Matching         & 10000  \\ 
flare-zh-afqmc      & Semantic Matching    & Semantic Matching         & 4316   \\ 
fincuge-finqa       & Financial Question Answering    & Question Answering         & 19906  \\ 
fincuge-fincqa      & Financial Question Answering    & Causal Question Answering       & 21965  \\ 
Duxiaoman/FinanceIQ & Financial Examination Multiple-Choice Questions & Single-Choice Questions        & 3573   \\ 
fingpt-fineval      & Financial Examination Multiple-Choice Questions & Single-Choice Questions        & 1056   \\ 
Duxiaoman/FinCorpus & Financial Examination Multiple-Choice Questions & Multiple-Choice Questions        & 40000 (filtered)  \\
\hline
\end{tabular}%
}
\end{table}

\textbf{Sentiment Analysis}

Sentiment analysis reflects the emotional trends in financial news, which can influence decision-making. We have collected four datasets, including flare-zh-fe, flare-zh-stocka, flare-zh-stockb, and CFSC-ABSA. The flare-zh-stocka dataset analyzes market data and company-related announcements, determining whether the company's stock movement is outperforming the market, neutral, or underperforming based on the impact of news on stock data. The remaining datasets involve judging the sentiment of sentences in given financial news, providing answers of positive, negative, or neutral sentiment.

\textbf{Text Classification}

For financial news classification, we selected the datasets ficuge-finnl, flare-zh-nl, and flare-zh-nl2. These datasets output 2-3 category keywords based on the given financial texts. For financial negative news and its subject determination, the flare-zh-nsp dataset judges whether a given entity contains negative news based on financial news and entities.

\textbf{Text Generation}

The text generation task mainly involves generating summaries from long financial news texts, requiring them to be concise yet contain key information. The datasets used are fincuge-finna and flare-zh-na.

\textbf{Information Extraction}

Information extraction tasks include event type and entity recognition, relationship extraction, and named entity recognition. The datasets used are flare-zh-19ccks, flare-zh-20ccks, flare-zh-21ccks, flare-zh-22ccks, fincuge-finre, and flare-zh-ner. The first four datasets are for event type and entity recognition, analyzing given financial texts to determine all event types and their corresponding subjects. The relationship extraction dataset fincuge-finre selects the correct relationship for an entity pair from 44 possible relationships based on financial texts. The named entity recognition dataset flare-zh-ner extracts entities from financial texts.

\textbf{Semantic Matching}

For the semantic matching task, we selected two datasets: flare-zh-corpus and flare-zh-afqmc, which involve determining whether the semantics expressed in two financial texts are consistent.

\textbf{Financial Question Answering}

For general financial question answering, we use the fincuge-finqa dataset, which extracts event information from financial texts and answers questions in context. The financial causal dataset fincuge-fincqa also focuses on causal relationships in financial texts.

\textbf{Financial Examination Multiple-Choice Questions}

This includes single-choice questions from Duxiaoman and FinGPT covering various financial fields, as well as multiple-choice questions from Duxiaoman's financial corpus, covering knowledge and calculations in banking, securities, accounting, funds, futures, and other areas.

\subsection{Examples of Financial NLP Tasks}

\begin{figure*}[htbp]
    \centering
    \includegraphics[width=0.7\textwidth]{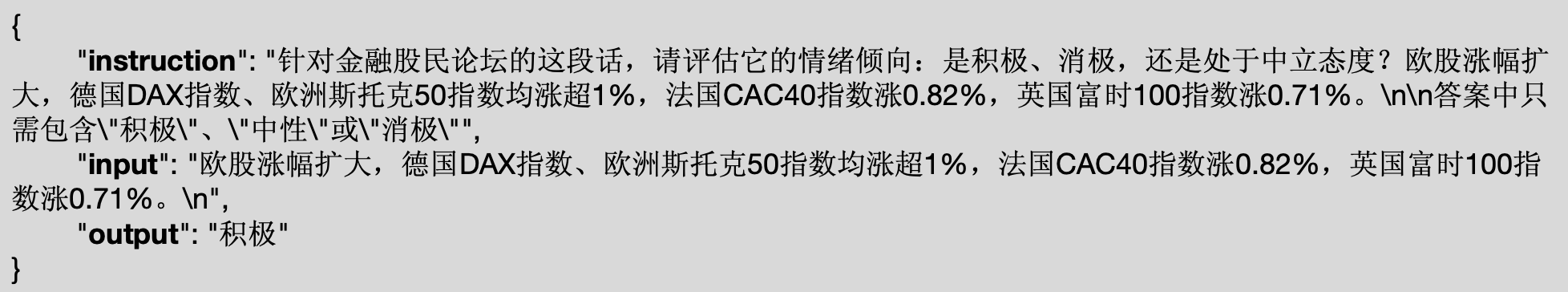}
    \caption{Example of Sentiment Analysis.}
    \label{img:sentiment_analysis}
\end{figure*}

\begin{figure*}[htbp]
    \centering
    \includegraphics[width=0.7\textwidth]{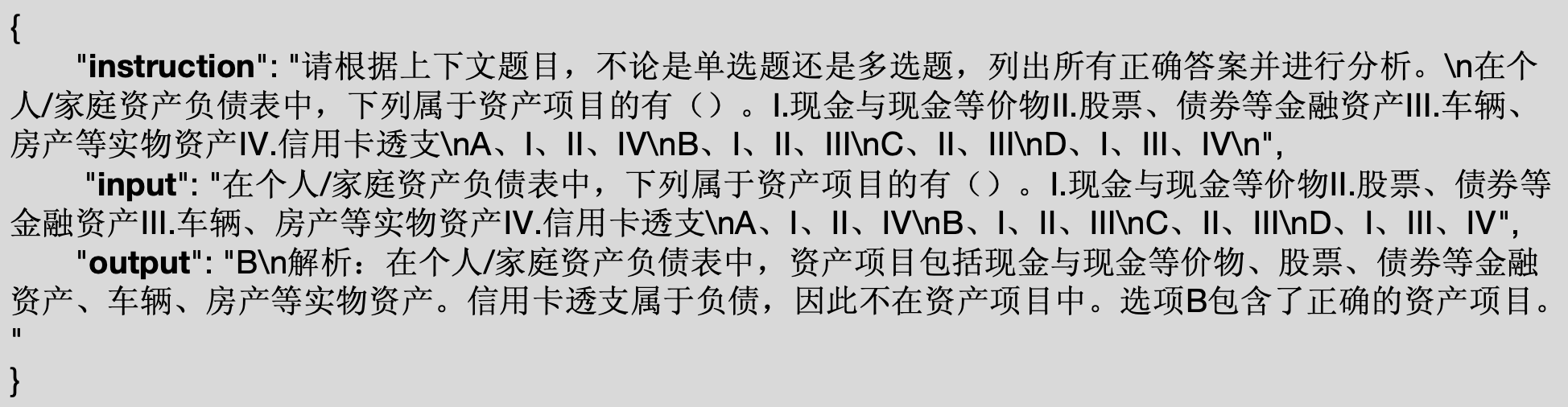}
    \caption{Example of Financial Examination Multiple-Choice Questions.}
    \label{img:financial_multi_choice}
\end{figure*}

All training data is unified into the triple \{instruction, input, output\} format. Specific examples of each type of dataset are shown in~\autoref{img:sentiment_analysis} to~\autoref{img:financial_multi_choice}.

\begin{figure*}[htbp]
    \centering
    \includegraphics[width=0.7\textwidth]{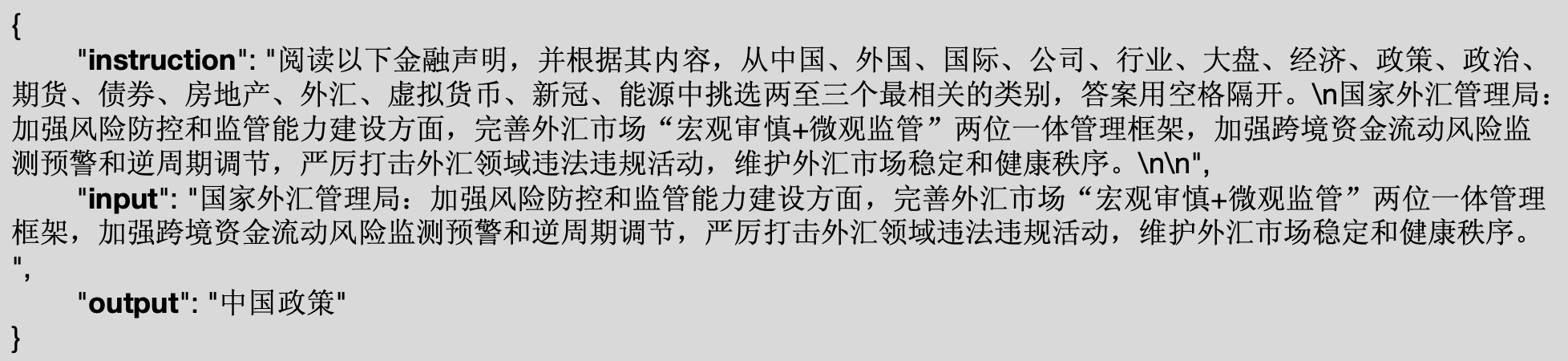}
    \caption{Example of Text Classification.}
\end{figure*}

\begin{figure*}[htbp]
    \centering
    \includegraphics[width=0.7\textwidth]{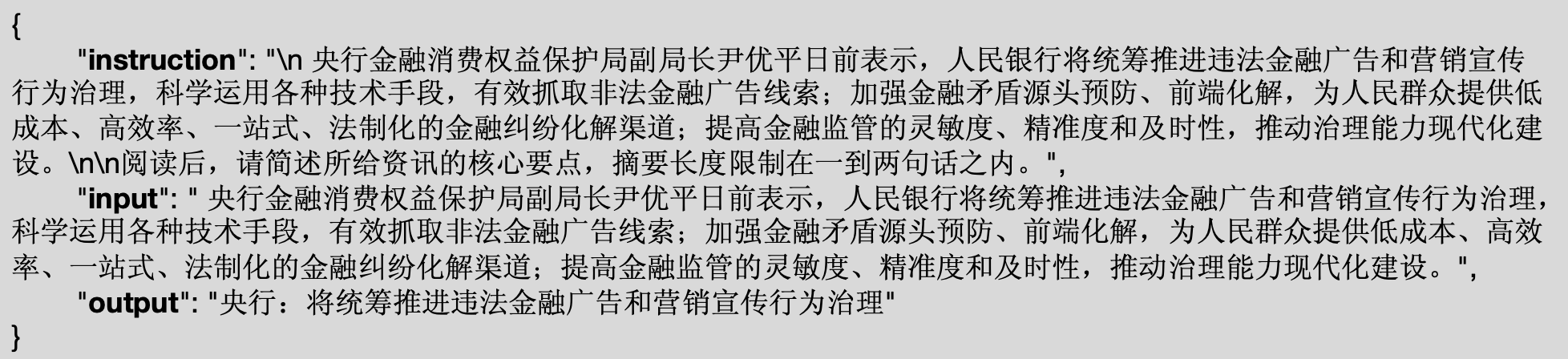}
    \caption{Example of Text Generation.}
\end{figure*}

\begin{figure*}[htbp]
    \centering
    \includegraphics[width=0.7\textwidth]{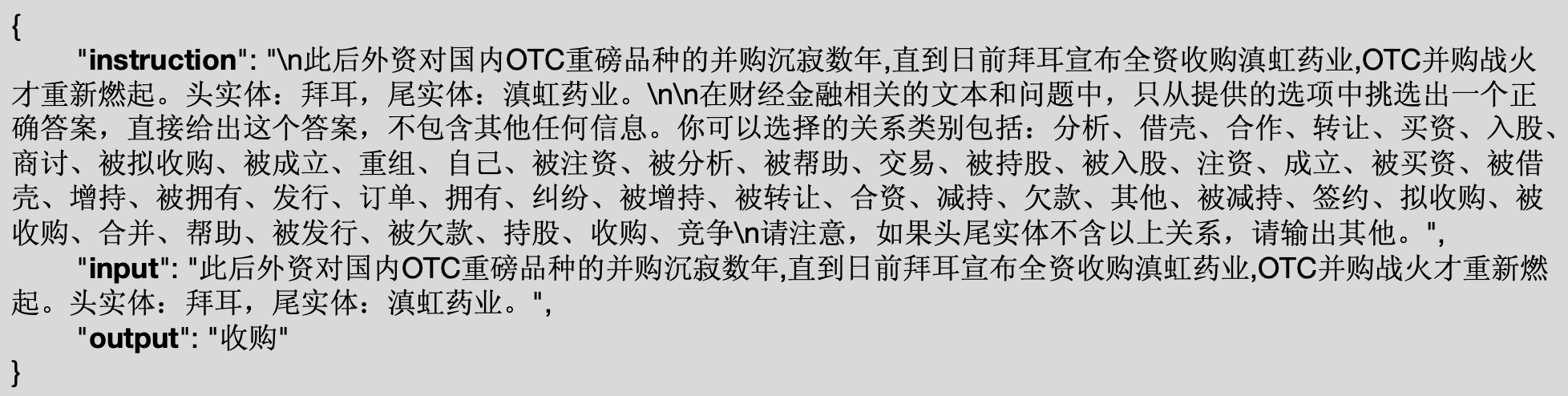}
    \caption{Example of Information Extraction.}
\end{figure*}

\begin{figure*}[htbp]
    \centering
    \includegraphics[width=0.7\textwidth]{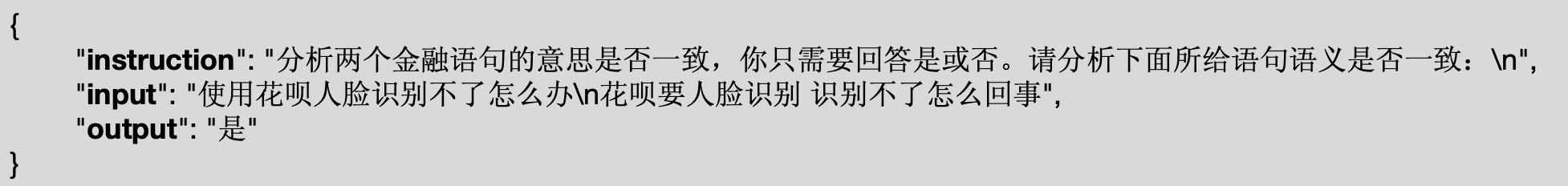}
    \caption{Example of Semantic Matching.}
\end{figure*}

\begin{figure*}[htbp]
    \centering
    \includegraphics[width=0.7\textwidth]{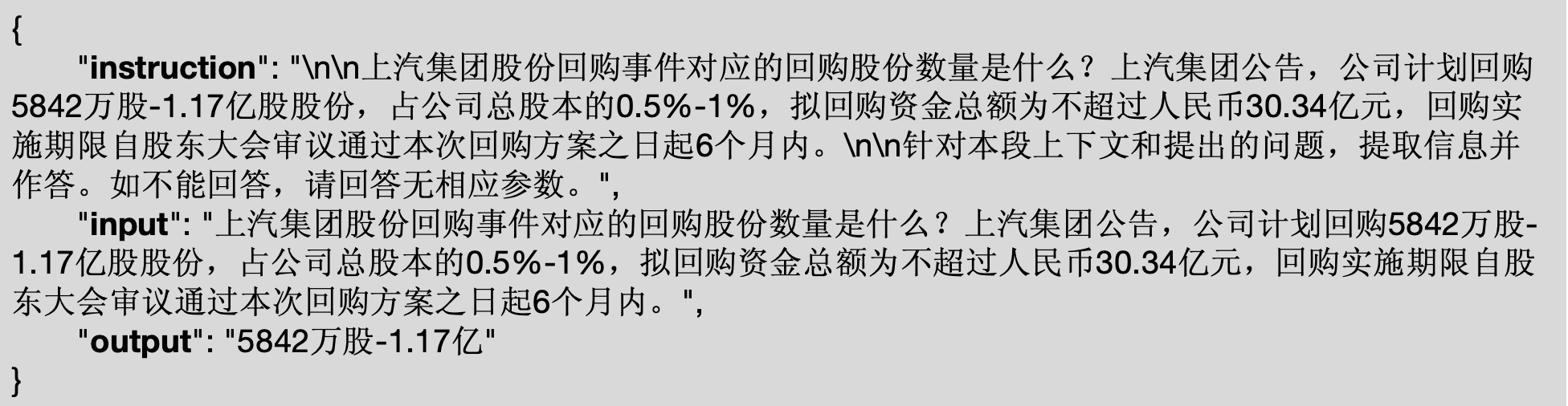}
    \caption{Example of Financial Question Answering.}
\end{figure*}

\subsection{Data Mixing Ratio Distribution}
\label{app:data_mix}
\begin{figure}
    \centering
    \includegraphics[width=0.7\textwidth]{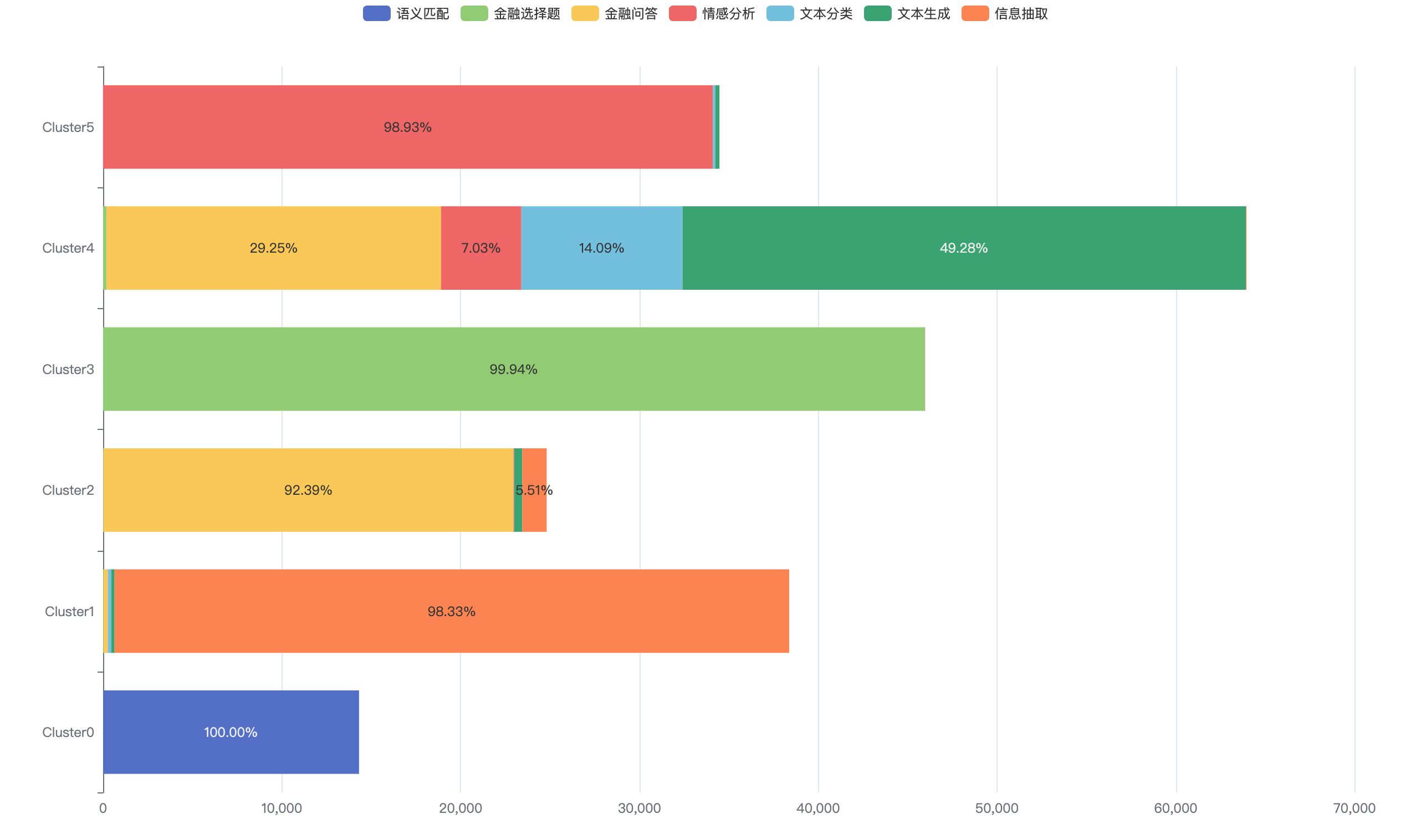}
    \caption{Data mixing ratio for each category after clustering.}
    \label{fig:all_sum}
\end{figure}
\begin{figure}
    \centering
    \includegraphics[width=0.7\textwidth]{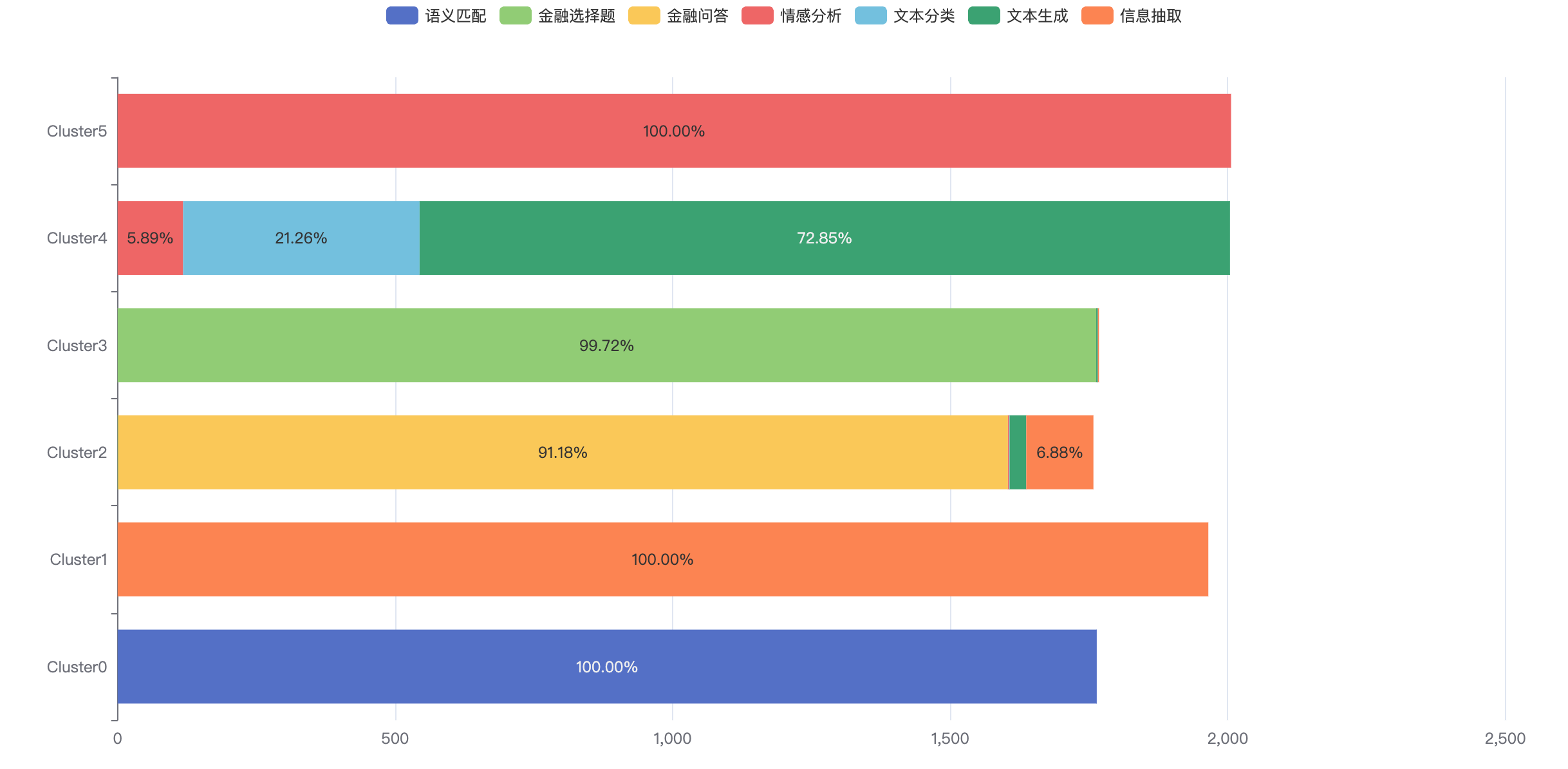}
    \caption{Data mixing ratio for each category after the first stage of data redistribution.}
    \label{fig:2k_sum}
\end{figure}
\begin{figure}
    \centering
    \includegraphics[width=0.7\textwidth]{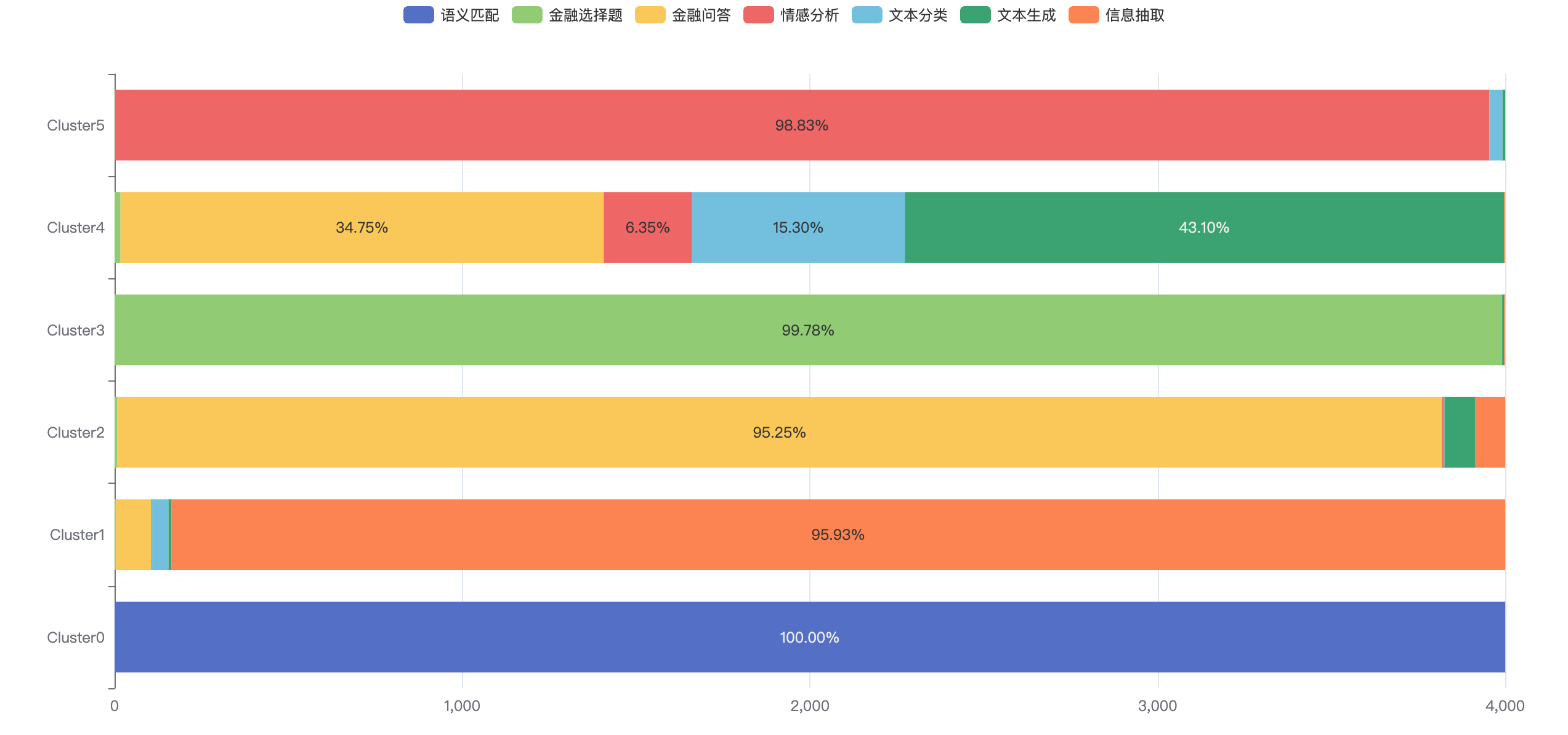}
    \caption{Data mixing ratio for each category after the second stage of data redistribution.}
    \label{fig:4k_sum}
\end{figure}

\subsection{A-DBSCAN}
\label{appendix:DBSCAN}

\begin{algorithm}[htbp]
\caption{Simplified Density-Based Spatial Clustering of Applications with Noise (A-DBSCAN) Algorithm with Sub-cluster Handling}
\DontPrintSemicolon  % 用于取消每行末尾的分号
\KwData{Dataset $D$, neighborhood size $k$}
\KwResult{Set of formed clusters $Clusters$}
\SetKwFunction{FHandleSubClusters}{HandleSubClusters} % 定义处理子簇的函数

Initialize cluster set $Clusters = \emptyset$\;
Compute the local density $\rho_p$ for each data point $p$: $\rho_p = \frac{1}{\frac{1}{k} \sum_{j=1}^{k} d(p, p_{j})}$\;
Sort the points in $D$ by $\rho$ to form a priority queue $Q$\;
Initialize $\varepsilon$ and $MinPts_{\text{init}}$: $\varepsilon = \text{median}\{d(p, p_{k})\} \forall p \in D$, $MinPts_{\text{init}} = \frac{\varepsilon \times \rho_{\text{max}}}{2}$\;
\While{Select the first point $p$ from $Q$}{
    \If{point $p$ has not been visited}{
        Mark $p$ as visited\;
        Get the $\varepsilon$-neighborhood of $p$, $N_p$\;
        \If{$|N_p| \geq MinPts_{\text{init}}$}{
            Initialize a new cluster $cluster$ and add $p$\;
            \For{each point $q \in N_p$}{
                \If{point $q$ has not been visited}{
                    Mark $q$ as visited\;
                    Get the $\varepsilon$-neighborhood of $q$, $N_q$\;
                    \If{$|N_q| \geq MinPts_{\text{init}}$}{
                        Add points in $N_q$ to $N_p$\;
                    }
                }
                \If{point $q$ does not belong to any cluster}{
                    Add $q$ to $cluster$\;
                }
            }
            $Clusters = Clusters \cup \{cluster\}$\;
            Update $MinPts$: $MinPts = \max\left(2, \frac{\rho_{p}}{\rho_{\text{max}}} \times MinPts_{\text{init}}\right)$\;
        }
    }
}

\FHandleSubClusters{$Clusters$}\;  % 调用处理子簇的函数

\SetKwProg{Pn}{Procedure}{:}{\KwRet} % 定义Procedure环境
\Pn{\FHandleSubClusters{$Clusters$}}{
    Compute the average quantity $N_{avg} = \frac{\sum_{i}N_{C_i}}{|Clusters|}$, where $N_{C_i}$ is the number of data points in sub-cluster $C_i$\;
    
    \ForEach{sub-cluster $C_i \in Clusters$}{
        \uIf{$N_{C_i} > N_{avg}$}{
            Sampling: Randomly select $N_{avg}$ data points from $C_i$\;
        }
    }
}
\end{algorithm}

\end{document}